\documentclass{article}

\usepackage[preprint]{icml2020}  
\usepackage{times}
\usepackage{epsfig}
\usepackage{graphicx}
\usepackage{amsmath}
\usepackage{amssymb}
\usepackage{siunitx}
\usepackage{nicefrac}
\usepackage{multirow}

\usepackage{booktabs}       
\usepackage[T1]{fontenc}  
\usepackage{hyperref}

\usepackage[capitalize]{cleveref}

\usepackage{xspace}
\makeatletter
\DeclareRobustCommand\onedot{\futurelet\@let@token\@onedot}
\def\@onedot{\ifx\@let@token.\else.\null\fi\xspace}
\def\eg{\emph{e.g}\onedot} 
\def\ie{\emph{i.e}\onedot}

\def\etal{\emph{et al}\onedot}
\makeatother

\def\modelname{ObSuRF}
\def\shapenetname{MultiShapeNet}
\def\x{\mathbf{x}}
\def\xo{\mathbf{x_0}}
\def\d{\mathbf{d}}
\def\c{\operatorname{\mathbf{c}}}
\def\p{\operatorname{p}}
\def\r{\operatorname{\mathbf{r}}}
\def\q{\operatorname{q}}

\def\tfar{t_\text{far}}

\DeclareMathOperator*{\argmax}{arg\,max}
\Crefname{equation}{Eq.}{Eqs.}

\begin{document}

\twocolumn[
\icmltitle{Decomposing 3D Scenes into Objects via Unsupervised Volume Segmentation}



\icmlsetsymbol{equal}{*}

\begin{icmlauthorlist}
\icmlauthor{Karl Stelzner}{tuda}
\icmlauthor{Kristian Kersting}{tuda}
\icmlauthor{Adam R.\ Kosiorek}{dm}
\end{icmlauthorlist}

\icmlaffiliation{dm}{DeepMind, London}
\icmlaffiliation{tuda}{TU Darmstadt}
\icmlcorrespondingauthor{Karl Stelzner}{\texttt{stelzner@cs.tu-darmstadt.de}}

\icmlkeywords{Machine Learning, Unsupervised Learning, 3D, Scene Understanding, NeRF}

\vskip 0.3in
]



\printAffiliationsAndNotice{}  

\begin{abstract}
We present \modelname, a method which turns a single image of a scene into a 3D model
represented as a set of Neural Radiance Fields (NeRFs), with each NeRF corresponding to a different object. 
A single forward pass of an encoder network outputs a set of latent vectors describing the objects in the scene.
These vectors are used independently to condition a NeRF decoder, defining
the geometry and appearance of each object.
We make learning more computationally efficient by deriving a novel loss, which allows training
NeRFs on RGB-D inputs without explicit ray marching.
After confirming that the model performs equal or better than state of the art on three 2D image segmentation benchmarks,
we apply it to two multi-object 3D datasets: A multiview version of CLEVR, and a novel dataset in which scenes are populated by ShapeNet models.
We find that after training \modelname{} on RGB-D views of training scenes, it is capable of not only recovering the 3D geometry of a scene depicted in a single input image, but also to segment it into objects, despite receiving no supervision in that regard.

\end{abstract}


\section{Introduction}
The ability to recognize and reason about 3D geometry is key to a wide variety of important robotics and
AI tasks, such as dynamics modelling, rapid physical inference, robot grasping, or autonomous driving \cite{battaglia2013pnas,chenKZMFU16,mahler2019learning,20-driess-RSS,liMZFT020}.
Despite consistent efforts \cite{wang2018pixel2mesh, qi2017pointnet}, traditional representations
of geometry have proven difficult to integrate with continuous machine learning systems,
since they either rely on discrete structures (polygonal meshes) or exhibit undesirable 
scaling behavior (voxels, points clouds).

Recent work has revealed a potentially viable alternative: instead of representing shapes
explicitly, one may use a neural network to map 3D coordinates to binary occupancy
indicators \cite{mescheder2019occupancy}, signed distances to the shape \cite{park2019deepsdf}, or
volume radiances \cite{mildenhall2020nerf}. Such functions, when used together with an encoder network,
can facilitate learning low-dimensional \emph{implicit} representations of geometry \cite{mescheder2019occupancy, nerfvae}.
These representations are continuous and independent of resolution or scene dimensions, addressing the main downsides of
explicit alternatives.
Most recent work on neural radiance fields had focused on using them to render high quality images, and the ability to obtain
such fields from minimal supervision (such as a small set of RGB images) \cite{martinbrualla2021nerf, nerfvae, guo2020objectcentric}.
\begin{figure}
    \centering
    \includegraphics[width=\linewidth]{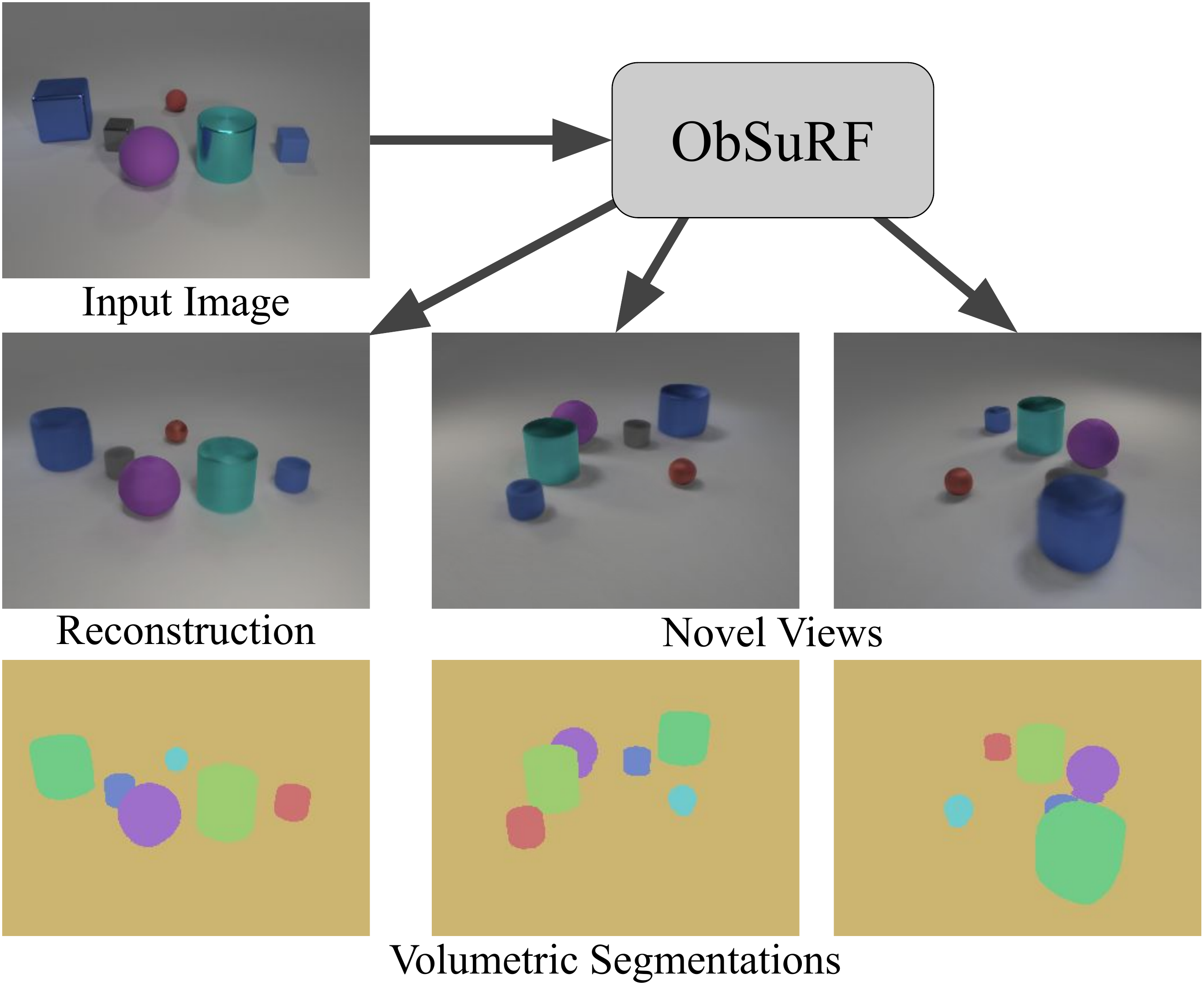}
    \caption{\modelname~ uses a single image (upper left corner) to infer a set of NeRF scene functions representing different objects.
    Since NeRFs represents 3D geometry and appearance, the scene and can then be rendered from arbitrary angles (middle row). 
    Despite receiving no supervision on how to segment scenes, \modelname{} learns to separate objects
    into individual NeRF components (bottom row).
    For a full demo, please watch the supplementary videos at \url{https://stelzner.github.io/obsurf/}.}
    \label{fig:eyecatcher}
\end{figure}

In this paper, we investigate a different aspect.
We ask whether implicit representations, when integrated in a latent variable model,
can be used to extract semantically meaningful information out of a given scene without supervision.
Such representations can then be used effectively for downstream reasoning tasks.
In particular, we focus on obtaining object-based
representations from multi-object scenes in an unsupervised way. Representations factored into objects are beneficial for dynamics modelling, visual question answering, and many other tasks
\cite{battaglia2016interaction, santoro2017relational, battaglia2018relational}.
Previous works in unsupervised object-based representation learning have mostly focused on segmenting 2D images \cite{eslami2016air, greff2020multiobject, locatello2020slotatt}.
A lack of suitable inductive biases has limited such methods to visually simple synthetic data, whereas more
realistic scenes with complex textures and geometries have remained out of reach \cite{weis2020unmasking}. 

To move towards more complex scenarios, we present \emph{\modelname},
a model which learns to decompose scenes consisting of multiple \emph{Objects} into a \emph{Superposition
of Radiance Fields}.
To achieve this, we encode the input image with a slot-based encoder
similar to \cite{locatello2020slotatt}, but use the resulting set of latent codes to condition continuous 3D scene functions
\cite{mildenhall2020nerf} instead of explicitly generating 2D images.
For training the model in 3D, we provide three \mbox{RGB-D} views of each training scene,
and optimize the model to match the observed
depths and colors. To do so, we reframe NeRF's volumetric rendering as a Poisson process
and derive a novel training objective, which allows for more efficient training
when depth information is available as supervision.
After confirming that the resulting model performs as well or better as previous
approaches on 2D image datasets, we test it on two new 3D benchmarks: A 3D version of CLEVR \cite{johnson2017clevr} featuring multiple viewpoints,
camera positions, and depth information, and \shapenetname{}, a novel multiobject dataset in which the objects are shapes from the ShapeNet dataset \cite{shapenet2015}.
We will publish both of these benchmarks and the code for our model with this paper.



\section{Related Work}
\modelname{} builds upon prior work on implicit representations of 3D geometry.
It is also related to previous approaches on unsupervised segmentation in both 2D and 3D, as well as set prediction.
\vspace{-1em}
\paragraph{Learned Implicit Representations of 3D Shapes.}
Early work focused on predicting the geometry of shapes using occupancy fields \cite{mescheder2019occupancy} or signed distance functions (SDFs) \cite{park2019deepsdf}. 
These models were trained using ground-truth geometry, for instance by sampling points from the meshes in the
ShapeNet dataset \cite{shapenet2015}.
Mildenhall \etal subsequently introduced Neural Radiance Fields (NeRF) \cite{mildenhall2020nerf},
which predict density and color of points in a scene.
This allows for differentiable rendering by tracing rays
through the scene and integrating over them.
The required supervision is thereby reduced to a small set of posed images.
However, explicit integration also makes this approach very computationally expensive.
Therefore, we opt for a pragmatic middle ground: we assume access to views of the training scenes
enriched with depth information (RGB-D), and show how this allows for training NeRFs without expensive raymarching.
We note that unlike ground-truth meshes, such data may be obtained at scale in the real world with reasonable accuracy
using either low-cost camera systems \cite{horaud2016depthcameras} or by estimating depth from stereo vision or motion, see \eg \cite{chang2018pyramid,Teed2020DeepV2D}.

 In its original form, NeRF is optimized to fit a single scene, which makes it too slow for real-time applications, and also 
 ill-suited for representation learning, as the scene is encoded in hundreds of thousands of neural network weights.
 Very recent work has employed encoder networks to obtain latent representations of input scenes,
 which can then be used to condition a decoder shared between scenes \cite{grf2020, nerfvae}.
 We take a similar approach, but put a greater emphasis on compact and factored representations.

\vspace{-1em}
\paragraph{Unsupervised Image Segmentation.}
The work in this area can be broadly grouped in two categories.
Patch-based approaches represent scenes as a collection of object bounding boxes
\cite{eslami2016air, kosiorek2018squair, stelzner2019supair, crawford2019spatially}.
This encouraging spatial consistency of objects, but limits the model's ability to explain complicated objects.
Scene-mixture models represent scenes as pixelwise mixture models, with each mixture component corresponding to a single object
\cite{greff2016tagger, greff2017nem, steenkiste2018relational, burgess2019monet, greff2020multiobject, locatello2020slotatt, Engelcke2020GENESIS}. 
Since spatial consistency is not enforced, these models sometimes segment images by color rather than by object \cite{weis2020unmasking},
motivating combinations of the two concepts \cite{Lin2020SPACE}.
By employing a NeRF-based decoder (instead of a CNN), we keep the flexibility of pixelwise mixture models, but
mitigate their issues:
since NeRF is explicitly formulated as a function of spatial coordinates in a Fourier basis,
it is strongly biased toward spatial coherence.
We note that not a single model in this line of research has been shown to work well on natural images,
instead they have targeted synthetic benchmarks such as CLEVR \cite{johnson2017clevr}.

\vspace{-1em}
\paragraph{Unsupervised 3D Segmentation.} Most work in this area focuses on segmenting
single shapes into parts, and is trained on ground truth geometry.
BAE-NET \cite{chen2019bae_net} reconstructs voxel inputs as a union of occupancy functions.
CvxNet \cite{deng2020cvxnet} and BSP-Net \cite{chen2020bspnet} represent shapes as a union of convex polytopes,
with the latter capable of segmenting them into meaningful parts.
Similarly, UCSG-Net \cite{kania2020ucsgnet} combines SDF primitives via boolean logic operations,
yielding rigid but interpretable shape descriptions.

Only a small number of publications attempt to segment full scenes into objects. Ost \etal 
train a model to segment a complex real-world video into objects,
but doing so requires a full scene graph, including manually annotated tracking information such as
object positions \cite{ost2021neural}.
In GIRAFFE \cite{niemeyer2020giraffe}, object-centric representations emerge when a set of NeRFs conditioned
on latent codes is trained as a GAN.
However, the focus lies on image
\emph{synthesis} and not \emph{inference}. Consequently, it is not straightforward to obtain representations for a given scene.
BlockGAN \cite{BlockGAN2020} achieves similar results, but uses latent-space perspective projections and CNNs instead of NeRFs.
A recent paper by Elich \etal is perhaps the closest to our work \cite{elich2021semisupervised}.
They encode a single input image into a set of deep SDFs representing the objects in the scene.
In contrast to our work, theirs requires pretraining on ground-truth shapes and operates on simpler scenes without shadows or reflections.


\vspace{-1em}
\paragraph{Set Prediction.} 
Objects present in a scene are interchangeable and form a set. 
Any attempt to assign a fixed ordering to the objects (such as left-to-right) will introduce discontinuities as they switch positions
(\eg as they pass behind each other) \cite{Zhang2020FSPool}.
It is therefore essential to use \emph{permutation equivariant} prediction methods in such settings.
We build on recent works which construct permutation equivariant set
prediction architectures \cite{locatello2020slotatt, kosiorek2020conditional, stelzner2020generative}.

\begin{figure*}
    \centering
    \includegraphics[width=\textwidth]{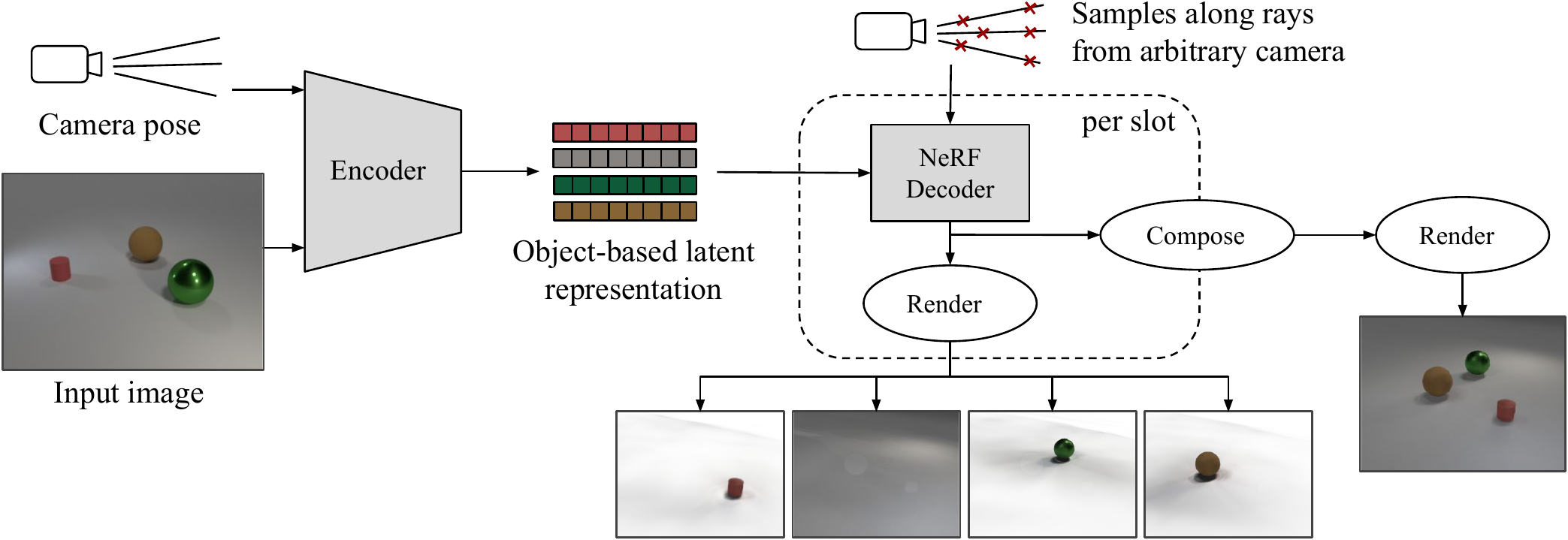}
    \caption{\modelname{} architecture.
    The encoder is given an input image with the
    corresponding camera pose, and infers an object-based latent representation 
    consisting of slots. These slots are used independently to condition a shared NeRF decoders.
    The volumes represented by the resulting scene functions may be rendered individually by 
    querying them along rays coming from an arbitrary camera. Alternatively, the volumes may be
    composed in order to render the full scene.
    }
    \label{fig:arch}
\end{figure*}

\section{Methods}
\modelname{} decomposes 3D scenes into objects, with each object modelled as a NeRF.
This requires evaluating multiple scene functions when rendering an image,
leading to a higher computational cost (by a factor of $n$ for $n$ NeRFs).
To make training more efficient, we derive a novel loss for training NeRFs on RGB-D data
based on a probabilistic view of the ray marching integral.
Additionally, this allows us to find a principled way of composing multiple NeRFs.
With these building blocks, we present the overall \modelname{} model.

\subsection{Neural Radiance Fields (NeRFs)}
\label{sec:nerf-intro}
NeRFs \cite{mildenhall2020nerf} represent the geometry and appearance of a scene as a neural network $f: (\x, \d) \rightarrow (\c, \sigma)$ 
mapping world coordinates $\x$ and viewing direction $\d$ to a color value $\c$ and a density value $\sigma$.
To guarantee coherent geometry, the architecture is chosen such that the density $\sigma$ is independent of the viewing direction $\d$.
To simplify notation, we refer to the individual outputs of $f(\cdot)$ as $\sigma(\x)$ and $\c(\x, \d)$.
NeRFs allow rendering images using classic ray marching techniques for volume rendering \cite{blinn1982light, kajiya1984ray}.
Specifically, the color $\hat{C}(\r)$ corresponding to the ray $\r$ is given by
\begin{equation}
    \hat{C}(\r) = \int_0^{\infty} T(t) \sigma(\r(t)) c(\r(t), \d) dt\,,
    \label{eq:volume_rendering}
\end{equation}
with transmittance $T(t) = \exp{(-\int_0^t  \sigma(\r(t')) dt') } $.
Since this procedure is differentiable, NeRFs are trained by minimizing the L2 loss between the rendered colors
$\hat{C}(\r)$ and the colors of the training image $C(\r)$, \ie $\mathcal{L}_\text{NeRF}(\r) = \|\hat{C}(\r) - C(\r)\|_2^2 $.
This is expensive, however, as approximating \Cref{eq:volume_rendering} requires evaluating $f$ many times per pixel (256 in Mildenhall \etal).
We will show that, if depth is available, we can instead train NeRFs with just two evaluations per pixel,
reducing the required computation by a factor of 128 per iteration.

\subsection{Volume Rendering as a Poisson Process}
While ray marching in \Cref{eq:volume_rendering} allows computing the ray color, we are also interested in finding the distribution of distances that give rise to this color---this will allow using depths in RGB-D data for supervision.
To this end, we show that \Cref{eq:volume_rendering} derives from an \emph{inhomogenous spatial Poisson process} \cite{blinn1982light,moller2003statistical}.



Consider a ray $\r(t) = \xo  + \d t$ traveling through a camera at position $\xo$
along direction $\d$.
The probability that light originating at point $\r(t)$ will not scatter and reach the camera unimpeded (transmittance)
is equal to the probability that no events occur in the spatial Poisson process $T(t) = \exp{(-\int_0^t  \sigma(\r(t')) dt') } $.
NeRF does not model light sources explicitly, but assumes that lighting conditions are expressed
by the color value of each point.
As a result, the light emitted at the point $\r(t)$ in the direction
of the camera $-\d$ has the color $\c(\r(t), \d)$, and its intensity is
proportional to the density of particles $\sigma(\r(t))$ at that point.
Consequently, the amount of light \emph{reaching} the camera from $\r(t)$ is proportional to
\begin{equation}
    \p(t) = \sigma(\r(t))T(t)\,.
    \label{eq:depthdist}
\end{equation}
In fact, under mild assumptions, $\p(t)$ is exactly equal to the distribution of possible depths
$t$ at which the observed colors originated\footnote{
Note that the transmittance is symmetric, and that both occlusion and radiance depend on the density $\sigma(\cdot)$.
Therefore, if a light source is present at $\xo$, the distribution of points along $\r$ it illuminates is also given by \Cref{eq:depthdist}.
It can be convenient to think of ray marching as shooting rays of light from the camera into the scene instead of the other way around.}.
We provide the full derivation in \Cref{secapp:nerf_depth}.

We can now reframe \Cref{eq:volume_rendering} as the expected color value under the
depth distribution $p(t)$:
\begin{equation}
    \hat{C}(\r) = \mathbb{E}_{t \sim \p(\cdot)}[\c(\r(t), \d)] = \int_0^{\infty} \p(t) c(\r(t), \d) dt\,.
    \label{eq:meancolor}
\end{equation}
Typically, one chooses a maximum render distance $t_\text{far}$ as the upper bound for the integration interval.
This leaves the probability $\p(t > t_\text{far}) = T(\tfar)$ that light from beyond $\tfar$ is missed.
To account for that, \Cref{eq:meancolor} is renormalized after approximating it with samples $0 \leq t_i \leq \tfar$.

\subsection{Fast Training on Depth and Colors}
\label{sec:rgbd-losses}
We now turn to the optimization objectives we use to train \modelname{}. 
Even though \citet{mildenhall2020nerf} use a hierarchical sampling scheme, training NeRFs requires many function evaluations for each ray and is therefore extremely expensive.
Moreover, as we show in \Cref{secapp:nerf_biased}, this sampling produces a biased estimate for $\hat{C}(\r)$, and, consequently $\mathcal{L}_\text{NeRF}$.
This is caused by the nested integration in \Cref{eq:volume_rendering}: If too few samples are collected, there is a significant chance
that thin, high-density volumes are missed entirely, even though they would dominate the color term $\hat{C}(\r)$ if it were to
be evaluated analytically. See \eg \citet{rainforth2018Onested} for a thorough investigation of this type of issue.

To avoid this computational cost during training, we use depth supervision by training on RGB-D data, \ie images for which the distance $t$ between the camera and the visible surfaces is known. 
Instead of integrating over $t$, this allows us to directly maximize the depth log-likelihood $\log \p(t)$ in \Cref{eq:depthdist} for the known values of $t$.
While we still need to approximate the inner integral,
it can be easily estimated by sampling from a Uniform distribution $\q(\cdot) = \text{Uniform}(0, \tfar)$ without
introducing bias:
\begin{equation}
    \log \p(t) = \log \sigma(\r(t)) - t \mathbb{E}_{t' \sim \q(\cdot)} \left[\sigma(\r(t'))\right]\,.
    \label{eq:depthdist_uniform_sampling}
\end{equation}
Since $\sigma(\r(t'))$ is likely to be large near the surface $t$ and close to zero for $t' \ll t$, the variance of this estimator can be very large---especially at the beginning of training.
To reduce the variance, we importance sample $t'$ from a proposal $q'(\cdot)$ with a higher density near the end of the integration range,
\begin{equation}
    \log \p(t) = \log \sigma(\r(t)) - \mathbb{E}_{t' \sim q'(\cdot)} \left[ \sigma(\r(t'))/q'(t') \right]\,.
    \label{eq:depthdist_importance_sampling}
\end{equation}
In practice, we set $q'(\cdot)$ to be an even mixture of the uniform distributions from $0$ to $0.98t$ and from
$0.98t$ to $t$, \ie, we take $50\%$ of the samples from the last $2\%$ of the ray's length. 

We fit the geometry of the scene by maximizing the depth log-likelihood of \Cref{eq:depthdist_importance_sampling}.
Staying with our probabilistic view, we frame color fitting as maximizing the log-likelihood
under a Gaussian distribution.
Namely, since we know the point $\r(t)$ at which the incoming light originated, we can evaluate the color likelihood as
\begin{equation}
    \p(C\mid\r(t), \d) = \mathcal{N}(C \mid \c(\r(t), \d), \sigma_C^2)\,,
    \label{eq:color_dist}
\end{equation}
with fixed standard deviation $\sigma_C$.
Overall, evaluating the joint log-likelihood $\log \p(t, C)$
requires only two NeRF evaluations: At the surface $\r(t)$ and at some point $\r(t')$ between the camera and the surface.
In practice we take the sample for the surface at $\r(t + \epsilon)$ with $\epsilon \sim \text{Uniform}(0, \delta)$.
This encourages the model to learn volumes with at least depth $\delta$ instead of extremely thin, hard to render surfaces.

%
%
%
%
%
%

\subsection{Composing NeRFs}
\label{sec:compose-nerfs}
We are interested in segmenting scenes into objects by representing each of them as an independent NeRF.
A scene is then represented by a set of NeRFs $f_1, \dots, f_n$, each with densities $\sigma_i$ and colors $\c_i$.
We now show how to compose these NeRFs into a single scene function. We arrive at the same equations as
\cite{niemeyer2020giraffe, guo2020objectcentric, ost2021neural}, but derive them by treating 
the composition of NeRFs as the \emph{superposition} \cite{moller2003statistical} of independent Poisson point processes,
yielding a  probabilistic interpretation of the algorithm.

Specifically, we assume that, while we move along the ray $\r$, each of the $n$ Poisson processes has a chance of triggering an event independently.
The total accumulated transmittance $T(t)$ from $\r(t)$ to $\xo$---the probability of encountering no events along the
way---should therefore be equal to the product of the transmittances $T_i(t)$ of each process:
\begin{equation}
    T(t) = \prod_{i=1}^n T_i(t) = \exp \left( -\int_0^t \sum_{i=1}^n \sigma_i(\r(t)) dt'\right).
    \label{eq:transmittance_composition}
\end{equation}
This is equivalent to another Poisson process with density $\sigma(\x) = \sum_i \sigma_i(\x)$.

To compute the color value, we additionally need to determine to what degree each of the component NeRFs is responsible for the incoming light.
Following \Cref{eq:depthdist}, the probability that NeRF $i$ is responsible for the light reaching the camera
from depth $t$ is $\p(t, i) = \sigma_i(\r(t))T(t)$.
Similarly to \Cref{eq:meancolor}, we can compute the color of a pixel by marginalizing
over both depth $t$ and component $i$, yielding $\hat{C}(\r) =$
\begin{align}
    \mathbb{E}_{t, i \sim \p(\cdot)}[\c_i(\r(t), \d)] = \int_0^{\infty} \sum_{i=1}^n \p(t, i) c_i(\r(t), \d) dt.
    \label{eq:render_composition}
\end{align}
It can also be useful to marginalize only one of the two variables. Marginalizing $i$ yields the depth distribution $\p(t) = \sum_i \p(t, i)$
which we use to render scenes via hierarchical sampling as in \cite{mildenhall2020nerf}. By marginalizing $t$, one obtains
a categorical distribution over the components $\p(i) = \int_t \p(t, i)$, which we use to draw segmentation masks. Finally, in
order to compute the color loss derived above, we use the expected color $\c(\x, \d) = \sum_i \c_i(\x, \d)\sigma_i(\x)/\sigma(\x)$.

\subsection{Learning to Encode Scenes as Sets of NeRFs}
We now describe the full encoder-decoder architecture of \modelname{} (see \Cref{fig:arch}), which can
\emph{infer} scene representations from a single view of that scene---for \emph{any} scene from a large dataset.
\modelname{} then turns that representation into a set of NeRFs, which allows
rendering views from arbitrary viewpoints and provides a full volumetric segmentation of the scene.

The encoder network $f_\text{enc}$ infers a set of latent codes $z_1, \dots, z_n$ (called slots) from the input image and the associated camera pose.
Each slot represents a separate object, the background, or is left empty.
By using a slot $z_i$ to condition an instance of the decoder network $f_\text{dec}$, we obtain 
the component NeRF $f_i(\cdot, \cdot) = f_\text{dec}(\cdot, \cdot; z_i)$.
In practice, we set the number of slots $n$ 
to one plus the maximum number of objects per scene in a given
dataset to account for the background.
 
\vspace{-1em}
\paragraph{Encoder.}
Our encoder combines recent ideas on set prediction \cite{locatello2020slotatt, kosiorek2020conditional}.
We concatenate the pixels of the input image with the camera position $\xo$, and a positional encoding \cite{mildenhall2020nerf}
of the direction $\d$ of the corresponding rays.
Such prepared input is then encoded into a feature map of potentially smaller resolution.
We initialize the slots $z_i$ by sampling from a Gaussian distribution with learnable parameters.
Following \citet{locatello2020slotatt}, we apply a few iterations of cross-attention between the slots and the elements of the
feature map, interleaved with self-attention between the slots (similar to \citet{kosiorek2020conditional}).
These two phases allow the slots to take responsibility for explaining parts of the input, and facilitate better coordination between the slots, respectively.
The number of iterations is fixed, and all parameters are shared between iterations.
The resulting slots form the latent representation of the scene.
We note that it would be straightforward to use multiple views as input to the encoder,
but we have not tried that in this work.
For further details, we refer the reader to \Cref{secapp:enc-arch}.
\vspace{-1em}
\paragraph{Decoder.}
The decoder largely follows the MLP architecture of \citet{mildenhall2020nerf}: We pass the position-encoded inputs $\x$ through a series of fully-connected layers, eventually outputting the density $\sigma(\x)$ and a hidden code $\mathbf{h}$.
The color value $\c(\x, \d)$ is predicted using two additional layers from $\mathbf{h}$ and $\d$.
To condition the MLP on a latent code $z_i$, we use the code to shift and scale the activations at each hidden layer,
a technique which resembles AIN of \cite{dumoulin2017style, brock2018large}.
We explored using $z_i$ to predict an explicit linear coordinate transform between the object and the world space
as in \cite{niemeyer2020giraffe, elich2021semisupervised}, but have not found this to be beneficial for performance.
We also note that we deliberately choose to decode each slot $z_i$ independently.
This means we can treat objects as independent volumes, and in particular we can render one object at a time.
Computing interactions between slots during decoding using \eg attention (as proposed by \citet{nerfvae})
would likely improve the model's ability to output complex visual features such as reflections, but would
give up these benefits.
Further details are described in \Cref{secapp:dec-arch}.

\vspace{-1em}
\paragraph{Training.}
The model is optimized using a batch of scenes at every training iteration.
For every scene, we use one image (and its associated camera pose, but without depth) as input for the encoder.
This yields the latent state $\{z_i\}_{i=1}^n$.
We then compute the loss on a random subset of rays sampled from the available RGB-D views of that scene.
This loss is then averaged over rays and scenes in the batch.

One advantage of 3D representations is that they allow us to explicitly express the prior knowledge
that objects should not overlap. 
This is not possible in 2D, where one cannot distinguish
between occluding and intersecting objects.
We enforce this prior by adding the \emph{overlap} loss
\begin{equation}
    \mathcal{L}_O(\r) = \sum_i \sigma_i(\r(t)) -
    \max_i \sigma_i(\r(t)),
\end{equation}
optimizing the overall loss
$\mathcal{L} = -\log p(t, C) + k_O \mathcal{L}_O(\r)$.
Experimentally, we find that $\mathcal{L}_O$ can prevent the model from learning object geometry at all when present from the beginning.
We therefore start with $k_O=0$ and slowly increase its value in the initial phases of training.
In turn, we find that we do not need the learning rate warm-up of \citet{locatello2020slotatt}.
We describe all hyperparameters used in \Cref{secapp:hyperparameters}.

\section{Experimental Evaluation}
\modelname{} is designed for unsupervised volumetric segmentation of multi-object 3D scenes.
As there are no published baselines for this setting, we start by evaluating it on 2D images instead of 3D scenes.
This allows us to gauge how inductive biases present in a NeRF-based decoder affect unsupervised segmentation.
We also check how a slotted representation affects reconstruction quality (compared to a monolithic baseline).
We then move on to the 3D setting, where we showcase our model on our novel 3D segmentation benchmarks; we also compare the reconstruction quality to the closest available baseline, NeRF-VAE \cite{nerfvae}.


\begin{table}[t]
    \centering
    \begin{tabular}{lccc}
    \toprule
         Model & Sprites (bin) & Sprites & CLEVR\\
         \midrule
         \modelname         & $\mathbf{74.4 \pm 1.8}$ & $\mathbf{92.4 \pm 1.3}$  & $98.3 \pm 0.8$  \\
        \hspace{0.9mm} sel. runs*         & $74.4 \pm 1.8$ & $93.1 \pm 0.3$ & $99.0 \pm 0.0$ \\
         SlotAtt. & $69.4 \pm 0.9$  & $91.3 \pm 0.3$ & $\mathbf{98.8 \pm 0.3}$ \\
         IODINE             & $64.8 \pm 17.2$ & $76.7 \pm 5.6$ & $\mathbf{98.8 \pm 0.0}$ \\
         MONet              & -               & $90.4 \pm 0.8$ & $96.2 \pm 0.6$ \\
         R-NEM              & $68.5 \pm 1.7 $ & -     & -     \\
         \bottomrule
    \end{tabular}
    \caption{Average foreground ARI on 2D datasets (in \%, mean $\pm$ standard deviation across 5 runs, the higher the better),
    compared with values reported in the literature \cite{locatello2020slotatt, greff2020multiobject, burgess2019monet, steenkiste2018relational}. Best values are bold. 
    (*) We notice that for one run on Sprites and 2 runs on CLEVR, \modelname{} learns to segment background from foreground.
    While that is generally desirable behavior, it slightly reduces the foreground ARI scores, since extracting
    exact outlines of the objects is more difficult. To quantify this effect, we also report results with these runs excluded. }
    \label{tab:2Dbenchmark}
\end{table}

\vspace{-1em}
\paragraph{Metrics.}
To evaluate segmentation quality in a way that is compatible with 2D methods,
we compute segmentations in 2D image space, and compare them to ground-truth using the \emph{Adjusted Rand Index} (ARI) \cite{rand1971objective, Hubert1985Comparing}.
The ARI measures clustering similarity and is normalized such that random segmentations result in a score of $0$, while perfect segmentations in a score of $1$.
In line with \citet{locatello2020slotatt}, we not only evaluate the full ARI, but also the ARI computed on
the foreground pixels (Fg-ARI; according to the ground-truth).
Note that achieving high Fg-ARI scores is much easier when the model is not attempting to segment the background,
\ie, to also get a high ARI score: This is because ignoring the background allows the model to segment the objects using rough outlines instead of sharp masks.
To measure the visual quality of our models' reconstructions, we report the mean squared error (MSE) between the rendered images and the corresponding test images. 
To test the quality of learned geometry in the 3D setting we also measure the MSE between the depths obtained via NeRF rendering and the ground-truth depths in the foreground (Fg-Depth-MSE).
We exclude the background as a concession to the NeRF-VAE baseline, as estimating the distance of the background in the kind of data we use can be difficult from visual cues alone.

\subsection{Unsupervised 2D Object Segmentation}
\label{sec:results2d}
We evaluate our model on three unsupervised image segmentation benchmarks from the
multi-object datasets repository \cite{multiobjectdatasets19}:
CLEVR, Multi-dSprites, and binarized Multi-dSprites. We compare against four recent models as state of the
art baselines: SlotAttention \cite{locatello2020slotatt}, IODINE \cite{greff2020multiobject},
MONet \cite{burgess2019monet}, and R-NEM \cite{steenkiste2018relational}.
We match the experimental protocol established
by \cite{locatello2020slotatt}: On dSprites, we train on the first 60k samples, on CLEVR, we select the first
70k scenes for training, and filter out all scenes with more than 6 objects. Following prior work, we test
on the first 320 scenes of each validation set \cite{locatello2020slotatt, greff2020multiobject}.
We also take the same image preprocessing steps on CLEVR as \cite{locatello2020slotatt, greff2020multiobject},
by cropping to a square section in the center and resizing to $64 \times 64$.

\begin{table}[t]
    \centering
    \begin{tabular}{lccc}
    \toprule
    Model                 & Sprites (bin) & Sprites & CLEVR \\
    \midrule
    \modelname{}          &  $1.34 \pm 0.15$ & $0.64 \pm 0.11$ & $0.41 \pm 0.06$\\
    NeRF-AE     & 4.82 & 2.60 & 1.75 \\
    \bottomrule
    \end{tabular}
    \caption{Testset reconstruction MSEs $\times 10^3$ of our model compared to the NeRF-AE ablation, which
    uses only a single latent vector.}
    \label{tab:2d-mses}
\end{table}

To adapt our encoder to the 2D case, we apply the positional
encoding introduced by \citet{locatello2020slotatt}, instead of providing camera position and ray directions.
To reconstruct 2D images, we query the decoder on a fixed grid of
2D points, and again drop the conditioning on viewing direction.
During training, we add a small amount of Gaussian noise to this grid to avoid overfitting to its
exact position.
Similar to previous methods \cite{locatello2020slotatt, greff2020multiobject}, we combine the 2D images which
the decoder produces for each slot by interpreting its density
outputs $\sigma_i$ as the weights of a mixture, \ie, we use the formula $\hat{C}(\r) = \sum_i  c_i(\r) {\sigma_i}/\sigma$.
As we show in \Cref{secapp:superposition_mixture},
this is equivalent to shooting a ray through a superposition of volumes with constant densities $\sigma_i$
and infinite depth. We train the model by optimizing the color reconstruction loss.

In \Cref{tab:2Dbenchmark}, we compare the foreground ARI scores achieved by our model with those from
the literature. We find that our model performs significantly better on the Sprites data, especially
the challenging binary variant, indicating that our decoder architecture imparts useful spatial priors.
On CLEVR, our model largely matches the already nearly perfect results obtained by prior models. 
We note that like \cite{locatello2020slotatt}, our model occasionally learns to segment the background in 
a separate slot, which is generally desirable but slightly reduces the Fg-ARI score as discussed above.

\begin{figure*}[t]
    \centering
    \includegraphics[width=\linewidth]{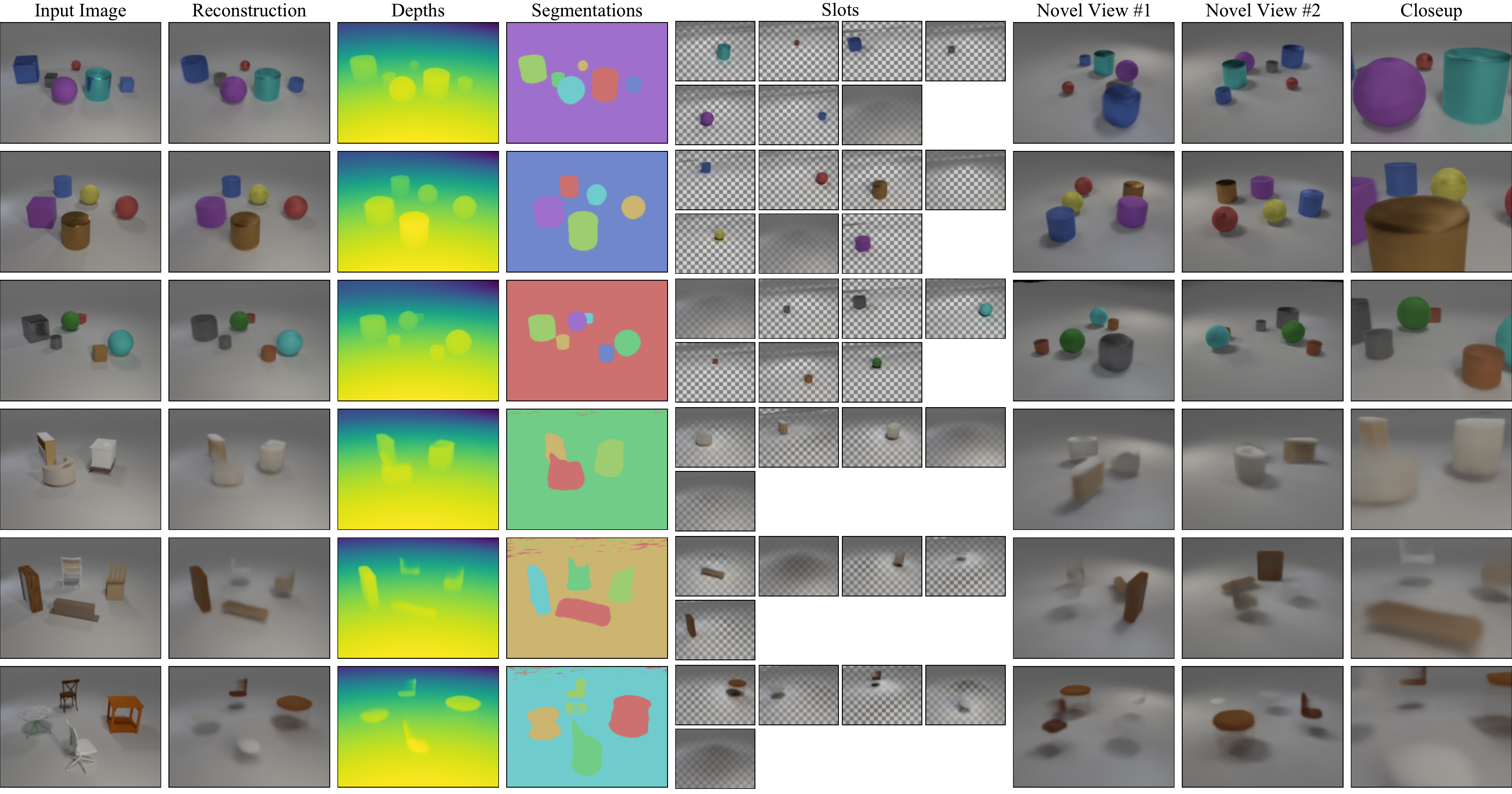}
    \caption{Qualitative results of applying \modelname{} to CLEVR-3D (top) and \shapenetname{} (bottom).
    We find that the model accurately models the geometry and appearance of CLEVR scenes, and decomposes
    them into objects. It also produces details which are not visible in the input image, such as
    the full shape of partially occluded objects, or the shadows on the backsides of objects.
    The \shapenetname{} dataset features a much larger variety of more complicated objects, such that
    \modelname{} cannot capture all of its details.
    However, it still learns to segment the scene into volumes corresponding
    to the objects, and to reproduce their dimensions, color, and position.
    Please see the supplementary video for a full demo.
    }
    \label{fig:samples}
\end{figure*}

To quantify the impact of learning representations which are segmented into slots, we construct a baseline
which uses only a single slot, called NeRF-AE. To keep the total computational capacity comparable to our model,
we set the size of this single latent vector to be equal to the total size of the slots \modelname{} is using.
We also double the size of the hidden layers in encoder and decoder. Still, as we show in Table
\ref{tab:2d-mses}, our model consistently achieves much lower reconstruction errors. This confirms previous
results indicating that representing multi-object scenes using a set of permutation equivariant elements
is not just helpful for segmentation, but also for prediction quality \cite{nerfvae, locatello2020slotatt}.
 
Training \modelname{} on these 2D datasets takes about 34 hours on a single V100 GPU. This is roughly 4 times faster
than SlotAttention (133 hours on one V100 \cite{locatello2020slotatt}) and almost 40 times more efficient
than IODINE (One week on 8 V100s \cite{greff2020multiobject}).
We attribute the efficiency gains compared to SlotAttention to our decoder architecture,
which processes each pixel individually via a small MLP instead of using convolutional layers.
 
\subsection{Unsupervised 3D Object Segmentation}
To test our model's capability of learning volumetric object segmentations in 3D, we assemble two novel benchmarks.
First, we adapt the CLEVR dataset \cite{johnson2017clevr} from \cite{multiobjectdatasets19} by rendering each scene
from three viewpoints, and also collecting depth values. The initial viewpoint is the same as in the original dataset,
the others are obtained by rotating the camera by $120^\circ$/$240^\circ$ around the $z$-axis. As in the 2D case, we restrict
ourselves to scenes with at most 6 objects in them.
Second, we construct \shapenetname{}, a much more challenging dataset which is structurally similar to CLEVR.
However, instead of simple geometric shapes, each scene is populated by 2-4 objects from the
ShapeNetV2 \cite{shapenet2015} 3D model dataset. To select shapes, we first uniformly choose between the
\emph{chairs}, \emph{tables} and \emph{cabinets} categories. We then select a random model from the
set of training objects in the chosen category,
and insert it into the scene. The resulting \shapenetname{} dataset contains $\num{11733}$ unique shapes.
To give an indication of the increased difficulty of segmenting this dataset into objects, we report
that the 2D version of our model achieves an ARI of $61.9\%$, and a Fg-ARI of $60.6\%$ on it.
We describe both datasets in detail in \Cref{secapp:data}. In contrast to the 2D case,
we utilize the full $320 \times 240$ resolution for both
datasets. Due to our decoder architecture, we do not need
to explicitly generate full size images during training,
which makes moving to higher resolution more feasible.

\begin{table*}[tp]
    \centering
    \begin{tabular}{lcccccccc}
    \toprule
         \multirow{2}{*}{Model} & \multicolumn{4}{c|}{CLEVR-3D} &\multicolumn{4}{c}{\shapenetname{}} \\
               & MSE $\times 10^3$& Fg-Depth-MSE & Fg-ARI & \multicolumn{1}{c|}{ARI} & MSE $\times 10^3$ & Fg-Depth-MSE & Fg-ARI & ARI  \\
         \midrule
         \modelname{}  & \textbf{0.78} & \textbf{0.10} & \textbf{95.7} & \textbf{94.6} & 1.81 & 3.44 & 81.4   & \textbf{64.1} \\
         \modelname{} w/o $\mathcal{L}_O$ & 0.80 & 0.12 & 85.5   & 4.83   & \textbf{1.78} & \textbf{3.40} & \textbf{94.4} & 16.5 \\
         NeRF-VAE & 4.7 & 1.03 & - & - &  5.5 & 110.3 & - & -  \\
         \bottomrule
    \end{tabular}
    \caption{Quantitative results on the 3D datasets. ARI scores are in percent, best values in bold.}
    \label{tab:3dresults}
\end{table*}

When testing our model, we provide a single RGB view of
a scene from the validation set to the encoder. We can then render the scene
from arbitrary viewpoints by estimating $\hat{C}(\r)$ (\Cref{eq:render_composition}) via hierarchical sampling
as in \cite{mildenhall2020nerf}. We set the maximum length
$\tfar$ of each ray to $40$ or the point at which it
intersects $z=-0.1$, whichever is smaller. This way,
we avoid querying the model underneath the ground plane,
where it has not been trained and where we cannot expect
sensible output. We use a rendering obtained from the
viewpoint of the input to compute the reconstruction MSE.
In order to obtain depthmaps and compute the depth-MSE,
we use the available samples to estimate the expected value of the
depth distribution $\mathbb{E}_{p(t)}[t] = \int_0^{\tfar} t \p(t) dt$.
We draw segmentation masks by computing $\argmax \p(i)$.
Finally, we also render individual slots by applying the same
techniques to the individual NeRFs, following \Cref{eq:depthdist,eq:meancolor}.
In order to highlight which space they do and do not occupy,
we use the probability $\p(t \leq \tfar)$ as an alpha channel,
and draw the slots in front of a checkerboard background.

We report results for two versions of our model:
The full version, which makes use of the overlap loss
$\mathcal{L}_O$ (\modelname{}), and an ablated version, which
does not (\modelname{} w/o $\mathcal{L}_O$). We choose the size of our slots to be 128 for CLEVR-3D and
256 for \shapenetname{}.
As a baseline, we compare to NeRF-VAE, which uses an autoencoder architecture similar to our model
to infer 3D geometry represented as a NeRF from a single 2D input image.
However, it is trained without depth information, and does
not attempt to segment the scene. Instead, it maximizes the evidence lower-bound (ELBO), comprised of a reconstruction term and a regularizng KL-divergence term. We note that we applied a small $\beta=10^{-3}$ to the KL term to improve reconstructions.
It also uses much larger latent representations of size $8 \times 8 \times 128$,
and an attention-based decoder.
While the differences in architecture make training times impossible to directly compare,
we estimate that due to our RGB-D based loss, \modelname{} required $24 \times$ fewer
evaluations of its scene function than NeRF-VAE over the course of training.

We present qualitative results obtained this way in
\Cref{fig:samples}, and quantitative results in \Cref{tab:3dresults}.
We find that \modelname{} learns to segment the scenes into objects, and to accurately reconstruct their position,
dimensions, and color. On the \shapenetname{} dataset, the large variety among objects causes the model to
sometimes output vague volumes instead of sharp geometry, whereas on CLEVR-3D, it stays close to the true scene.
We observe that similarly to the 2D models, the our model does not learn to segment the background into
its own slot without
$\mathcal{L}_O$. With this additional loss term however, it learns to also segment the background, leading to much
higher ARI scores. This somewhat diminishes the Fg-ARI score on \shapenetname{}, since, as discussed above,
doing well on both metrics is much harder. Compared to NeRF-VAE, we find that our model achieves much lower
reconstruction errors on both datasets. This is likely a result of the additional depth supervision our model is
receiving.

\subsection{Discussion}
With its results on the \shapenetname{} dataset, \modelname{} achieves unsupervised object segmentation in a
significantly more complex setting than previous models were able to \cite{locatello2020slotatt, weis2020unmasking}.
To achieve this, we have leveraged one of the most basic ideas in vision, namely the correspondence between 3D scene geometry
and 2D observations. In that sense, our model follows in the long tradition of vision-as-inverse-graphics approaches
(see \eg \cite{grenander76synthesis, grenander78analysis}). The main difference to previous methods in this direction
is that NeRFs provide us with a convenient method to parameterize arbitrary geometry and appearances in a fully continuous
way. This allows us to learn without prior knowledge on the shapes of objects in the dataset, in contrast to previous methods
which relied on fitting known meshes to the observed scene \cite{eslami2016air, romaszko2017vision}.

Compared with previous methods in 2D \cite{locatello2020slotatt},
\modelname{} has the advantage of observing 3D geometry through depth
information at training time. This provides it with a multitude of biases helpful for object segmentation
which are not present in the 2D setting: The spatial coherence of objects, the fact that they should not intersect,
and the effects of occlusion are all much more clearly expressed in the 3D setting. This allows us to move
beyond the limits of 2D segmentation outlined by \citet{weis2020unmasking}. While the question of how to
learn object-centric 3D representations from purely 2D observations is important scientifically, and recent results suggest
that it might be possible \cite{mildenhall2020nerf, nerfvae}, we conjecture that utilizing RGB-D data
will be an important accelerator for achieving unsupervised scene understanding in the real world.
A critical factor for this is the availability of large scale datasets, as current real-world multi-object
datasets gathered in controlled environments tend to not be bigger than around 100 scenes \cite{xiang2018posecnn,fang2020graspnet}.
         
\section{Conclusion}
We have presented \modelname{}, a model which learns to segment 3D scenes into objects represented as NeRFs.
We have shown that it learns to infer object positions, dimensions, and appearance on two challenging and
novel 3D modelling benchmarks. Importantly, it does so using purely observational data, without requiring 
supervision on the identity, geometry or position of individual objects.

Future work may extend our model in a variety of ways: Moving to a variational
architecture such as \cite{nerfvae} could help it better model uncertainties. Attending to multiple input views
or applying iterative inference \cite{grf2020, nerfvae} is likely to help scale to more challenging datasets.
Since our model learns to infer compact representations of the objects in a scene, using those representations
may be of interest for a variety of downstream tasks, including robot grasping, dynamics modelling, or visual
question answering. Finally, we reiterate the importance of gathering large datasets
in real but controlled environments, in order to enable
unsupervised scene understanding in the natural world.

\bibliographystyle{icml2020}
\bibliography{references}

\clearpage
\appendix
\section{Proofs and Derivations}
We start by providing detailed derivations for the results referenced in the main text.
\subsection{Derivation of the NeRF Depth Distribution}
\label{secapp:nerf_depth}
In \Cref{sec:nerf-intro}, we noted that the distribution $\p(t)$ shown in \Cref{eq:depthdist} arises
both as the distribution of points along $\r$ who contribute to the color
observed at $\xo = \r(0)$,
and as the distribution of points illuminated by a light source located at $\xo$.
Here, we start by showing the latter.

\subsubsection{Distribution of Light Originating at $\xo$}
Following the low albedo approximation due to \citet{blinn1982light}, we ignore indirect lighting.
In order for light originating at $\xo$ to illuminate $\r(t)$ directly, it must travel
along the ray without any scattering events. The rate of scattering events is given by
the inhomogenous spatial Poisson process with finite, non-negative density 
$\sigma(\x)$ \cite{moller2003statistical}.
We are therefore
looking for the distribution over the position of the first event encountered while
traveling along $\r$, also called the \emph{arrival} distribution of the process \cite{moller2003statistical}.
In a Poisson process, the probability of encountering no events along a given trajectory
$\r(t)$ with $t \geq 0$ (transmittance) is
\begin{equation}
    T(t) = \exp \left( \int_0^t -\sigma(\r(t')) dt' \right).
\end{equation}
The probability that light scatters before moving past some point $t_0$ is therefore
$\p(t \leq t_0) = 1 - T(t_0)$.
Let us now assume that $\lim_{t_0 \rightarrow \infty} T(t_0) = 0$, \ie, each ray of light
encounters a scattering event eventually. This is for instance guaranteed if the scene
has solid background. In that case, $1 - T(t)$ is the cumulative distribution function for
$\p(t)$, since it is non-decreasing, continuous, and fulfills $T(0) = 0, T(\infty) = 1$.
We can therefore recover the density function $\p(t)$ by differentiating:
\begin{align}
    \p(t) &= \frac{d}{dt} \left( 1- T(t) \right) \\
    &= - T(t) \frac{d}{dt} \int_0^t -\sigma(\r(t')) dt' \\
    &=\sigma(\r(t)) T(t).
    \label{eq:cdf_to_pdf}
\end{align}
\subsubsection{Distribution of Colors Observed at $\xo$}
As discussed in \Cref{sec:nerf-intro}, NeRF does not model light sources explicitly,
but assumes that lighting conditions are expressed by each point's RGB color value.
The observed color $\hat{C}(\r)$ at point $\xo$ along ray $\r$ is then a blend of the
color values along $\r(t)$ \cite{blinn1982light, kajiya1984ray}. 
The intensity with which we can observe the color emitted from point $\r(t)$ in direction $-\d$
of the camera is proportional to the density $\sigma(\r(t))$. Let $k$ denote the constant
of proportionality. Since $T(t)$ is the probability that light reaches $\xo$ from 
$\r(t)$, the we can observe the color $\c(\r(t), \d)$ at $\xo$ with intensity
$k\sigma(\r(t)) T(t)$. 

In order to obtain the distribution $\p(t)$ over the points from which
the colors we observe at $\xo$ originate, we must normalize this term by dividing by the
\emph{total} intensity along ray $\r$: $ \p(t) =$
\begin{equation}
    \frac{k\sigma(\r(t)) T(t)}{\int_0^\infty k\sigma(\r(t')) T(t') dt'} 
    = \frac{\sigma(\r(t)) T(t)}{\int_0^\infty \sigma(\r(t')) T(t') dt'}. 
\end{equation}
As we have shown in \Cref{eq:cdf_to_pdf}, the antiderivative of $\sigma(\r(t)) T(t)$
is $-T(t)$. The value of the improper integral is therefore
\begin{equation}
    \lim_{t \rightarrow \infty} \int_0^t \sigma(\r(t')) T(t') dt'
    = \lim_{t \rightarrow \infty}  T(0) - T(t) = 1.
\end{equation}
Consequently, $\p(t) = \sigma(\r(t))T(t)$. 
This procedure may be viewed as a continuous version of alpha compositing \cite{porter1984compositing}.

%

\subsection{Estimating $\mathcal{L}_\text{NeRF}$ is Biased}
\label{secapp:nerf_biased}

As a counterexample, consider the following situation in which a ray $\r$ passes through a thin but dense white volume,
before hitting black background. For simplicity,
we use grayscale color values.
\begin{align}
    \sigma(\r(t)) &=
        \begin{cases}
            100 & \text{if } 50 \leq t \leq 51 \\
            10 & \text{if } t > 80 \\
            0   & \text{otherwise},
        \end{cases} \\
    c(\r(t), \d) &=
        \begin{cases}
            1 & \text{if } 50 \leq t \leq 51 \\
            0 & \text{otherwise}.
        \end{cases}
\end{align}

We then have $T(50) = 1$, but $T(51) = T(80) = \exp(-100)$. Since $T(\cdot)$ is the CDF of $t$ (\cref{eq:cdf_to_pdf}),
this means almost all of the incoming light is coming from the thin volume at $t \in [50, 51]$:
\begin{align}
    p(50 \leq t \leq 51) &= 1 - \exp(-100) \\
    p(t > 80) &= \exp(-100).
\end{align}
Using \cref{eq:meancolor}, we find that the color value is almost exactly one:
\begin{equation}
    \hat{C}(\r) = 1 - \exp(-100) \approx 1.
\end{equation}
Let us now consider the result of estimating  $\hat{C}(\r)$ using $k=50$ stratified samples $t_1, \dots, t_k$, with
$\tfar = 100$. In this case, one sample $t_i$ is drawn uniformly from
each of $k$ evenly sized bins:
\begin{equation}
    t_i \sim \text{Uniform}\left(\frac{i-1 }{k}\tfar,\frac{i}{k} \tfar\right).
\end{equation}
Only one of the bins, the one for  $t_{26} \sim \text{Uniform}(50, 52)$, covers the volume
at $[50, 51]$. However, since this volume only makes up half the range of this bin, we have
$p(50 \leq t_{26} \leq 51) = 1/2$. Therefore, 50\% of the time, sampling in this way will result
in samples which miss the volume at $[50, 51]$ entirely. In those cases, we have
$c(\r(t_i), \d) = 0$ for all $i$.
As a result, the estimated color value is $\hat{C}^{|t_i}(\r) = 0$. 
Even if the estimated color is $1$ in all other cases, this sampling scheme will be a biased estimator for the true color values:
\begin{equation}
    \mathbb{E}_{t_i} \left[ \hat{C}^{|t_i}(\r) \right] \leq \frac{1}{2}  < \hat{C}(\r).
    \label{eq:nerf-biased}
\end{equation}
Collecting a second round of samples based on the estimated densities as proposed by Mildenhall \etal \cite{mildenhall2020nerf}
does not change the result. In the 50\% of cases where the thin volume was missed entirely during by the first set of samples,
we have $\sigma(\r(t_i)) = 0$ for all $t_i \leq 80$. As a result, the piecewise-constant PDF used for sampling the second round
of samples will equal $0$ for $50 \leq t \leq 51$, and none of the newly collected samples will cover the relevant range.
Therefore, \cref{eq:nerf-biased} also holds when the samples $t_i$ were collected using hierarchical sampling.

We note that this effect may be somewhat mitigated by the fact that NeRF models typically use different scene functions for
coarse and fine grained sampling \cite{mildenhall2020nerf, nerfvae}. The coarse scene function may then learn to model wider volumes
which are easier to discover while sampling, instead of the exact scene geometry. However, it seems clear that this will not
allow decreasing the number of samples arbitrarily, \eg to 2 as we did in this paper.

\subsection{Superposition of Constant Densities Yields a Mixture Model}
\label{secapp:superposition_mixture}
Here, we illustrate how a mixture model, like the one we use in \cref{sec:results2d},
naturally arises when a superposition of volumes with constant densities $\sigma_i$ and colors $c_i$ is rendered. 
Following \cref{eq:transmittance_composition} and writing $\sigma = \sum_i \sigma_i$, the transmittance along a ray $\r$ is
\begin{equation}
    T(t) = \exp\left(-\int_0^t \sigma dt' \right) = \exp(-\sigma t).
\end{equation}
As in \cref{sec:compose-nerfs}, we can then write
\begin{equation}
    p(t, i) = \sigma_i(\r(t)) T(t) = \sigma_i \exp(-\sigma t).
\end{equation}
Integrating over $t$ to obtain the distribution over the volumes $i$ yields
\begin{equation}
    p(i) = \int_0^\infty p(t, i) dt = \left[ \frac{\sigma_i}{-\sigma} \exp(-\sigma t) \right]_0^\infty = \frac{\sigma_i}{\sigma}.
\end{equation}
Using $c_i(\x) = c_i$, and following \cref{eq:render_composition}, we find that the observed color will then be 
\begin{equation}
    \mathbb{E}_{i \sim p(\cdot)} \left[ c_i \right] = \sum_{i=1}^n c_i \frac{\sigma_i}{\sigma},
\end{equation}
which is the desired mixture model.

\section{Architectures}
We now present the architectures used in our experiments. When we refer to some of the dimensions in terms of a variable, we specify the concrete
values used in \cref{secapp:hyperparameters}. \modelname{} uses exclusively ReLU activations.

\subsection{Encoder Architecture}
\label{secapp:enc-arch}
Due to the differences in resolution and complexity,
we used different feature extractor architectures for the 2D and 3D experiments.
The slot attention module was identical in both cases.

\vspace{-1em}
\paragraph{Feature Extractor for 2D Datasets.}
Following \cite{locatello2020slotatt}, we use a convolutional network with 4 layers
to encode the input image for the 2D datasets. Each layer has a kernel size of $5 \times 5$, and padding of $2$,
keeping the spatial dimensions of the feature maps constant at $64 \times 64$. Each layer outputs
a feature map with $d_h$ channels.
After the last layer, we apply the spatial encoding introduced by \cite{locatello2020slotatt}.
Again following \cite{locatello2020slotatt}, we process each pixel in the resulting feature map individually
using a layer normalization step, followed by two fully connected (FC) layers with output sizes $[d_h, d_z]$.

\vspace{-1em}
\paragraph{Feature Extractor for 3D Datasets.}
For the 3D experiments, we adapt the ResNet-18 architecture to achieve a higher model capacity.
We start by appending the camera position and a positional encoding \cite{mildenhall2020nerf} of the
viewing direction to each pixel of the input image, and increase the input size of the ResNet accordingly.
We remove all downsampling steps from the ResNet except for the first and the third, and keep the number of
output channels of each layer constant at $d_h$. Given an input image of size $240 \times 320$, the
extracted feature map therefore has spatial dimensions $60 \times 80$, and $d_h$ channels.
We apply the same pixelwise steps described for the 2D case (layer normalization followed by two FC layers)
to obtain the final feature map with $d_z$ channels.

\begin{table*}[t]
    \centering
    \begin{tabular}{llccc}
    \toprule
        Parameter & Description & 2D Experiments & CLEVR-3D & \shapenetname \\
        \midrule
        $d_z$ & Dimensionality of slots & 64 & 128 & 256 \\
        $d_h$ & Dimensionality of hidden layers & 64 & 128 & 256 \\
        $m$   & Number of Slot Attention iterations & 3 & 5 & 5 \\
        $n_f$ & Number of frequencies for pos. encoding & 8 & 16 & 16 \\
        $k_f$ & Exponent for lowest frequency & 0 & -5 & -5 \\
        $\sigma_\text{max}$ & Maximum density & - & 10 & 10 \\
        $\sigma_c$ & Standard deviation of color dist. & 1 & 0.2 & 0.2 \\
        $\delta$ & Noise added to depths $t$ & - & 0.07 & 0.07 \\
        $\hat{k}_O$ & Maximum overlap loss coefficient & - & 0.05 & 0.03 \\
        & Start of $\mathcal{L}_O$ ramp up & - & 50000 & 50000 \\
        & End of $\mathcal{L}_O$ ramp up & - & 100000 & 80000 \\
        & Batch Size & 128 & 64 & 64 \\
        & Rays per instance & - & 4096 & 2048 \\
        & Number of training iterations & 500k & 1000k & 500k\\
        \bottomrule
    \end{tabular}
    \caption{Hyperparameters used for the experiments with \modelname{}.}
    \label{tab:hyperparams}
\end{table*}

\vspace{-1em}
\paragraph{Slot Attention.}
We apply slot attention as described in \cite{locatello2020slotatt} with $n$ slots for $m$ iterations,
except for the following change. At each iteration, after the GRU layer, we insert a multi-head
self attention step with $4$ heads to facilitate better coordination between the slots.
Specifically, we update the current value of the slots via
\begin{equation}
    \mathbf{z}  := \mathbf{z} + \text{MultiHead}(\mathbf{z}, \mathbf{z}, \mathbf{z}).
\end{equation}
After $m$ iterations, the values of the slots form our latent representation $z_1, \dots z_n$.
We apply the same number of iterations $m$ during both training and testing.

\subsection{Decoder Architecture}
\label{secapp:dec-arch}
Here, we describe how density and colors are predicted from the spatial coordinates $\x$ and the
latent vector $z_i$. First, we encode the $\x$ via positional encoding \cite{mildenhall2020nerf},
using $n_f$ frequencies, with the lowest one equal to $2^{k_{f}} \pi$. We process the encoded positions
using a 5 layer MLP with hidden size $d_h$. After each hidden layer, we scale and shift the current hidden
vector $h$ elementwise based on $z_i$. Specifically, at each layer, we use a learned linear map to predict scale
and shift parameters
$\mathbf{\alpha}, \mathbf{\beta}$ (each of size $d_h$) from $z_i$, and update $\mathbf{h}$ via
$\mathbf{h} := (\mathbf{h} + \mathbf{\beta}) \cdot \mathbf{\alpha}$.

In the 2D case, we output 4 values at the last layer, namely the log density $\log \sigma$ and the RGB-coded
colors $\c$. In the 3D case, we output one value for the density, and $d_h$ values to condition the colors.
We append the view direction $\d$ to the latter, and predict the actual color output using two additional
FC layers of size $[d_h, 3]$, followed by a sigmoid activation. In the 3D case, we find that it is necessary
to set a maximum density value, as the model can otherwise arbitrarily inflate the depth likelihood $p(t)$
by increasing the density in regions which are known to always be opaque, such as the ground plane.
We do this by applying a sigmoid activation to the decoder's density output, and multiplying the result by
$\sigma_\text{max}$, obtaining values in the range $[0, \sigma_\text{max}]$.

\subsection{NeRF-AE Baseline Details}
For the NeRF-AE baseline (\cref{tab:2d-mses}), we adapt our encoder architecture to output a single vector instead
of a feature map. To this end, we first append a positional encoding \cite{mildenhall2020nerf} of each pixel's
coordinates to the input image.
This replaces the spatial encoding described above, since the method described by 
\cite{locatello2020slotatt} requires a full-sized feature map to be applicable.
We double the number of channels to $2d_h$, but set the convolutional layers to use
stride $2$, obtaining an output with spatial dimensions $4 \times 4$. As above, each of these vectors
is processed via layer normalization and two fully connected layers, this time with output sizes $[2d_h, 2d_h]$.
We flatten the result to obtain a vector of size $32 \cdot d_h$, and obtain the final vector using an MLP
of sizes $[4 d'_z, 2 d'_z, d'_z]$. This replaces the slot attention component described below. To keep the
total dimensionality of the latent space identical, we set $d'_z = n d_z$.

\subsection{NeRF-VAE Baseline Details}
\label{secapp:nerfvae}
We compare \modelname{} to NeRF-VAE on the quality of image reconstruction and depth estimation.
We use NeRF-VAE with the attentive scene function and without iterative inference as reported in \cite{nerfvae}, with the majority of parameters the same.
The only differences stem from different image resolution used in the current work ($240 \times 320$ as opposed to $64 \times 64$ in \cite{nerfvae}).
To accommodate bigger images, we use larger encoder with more downsampling steps.
Specifically, we use [(3, 2), (3, 2), (3, 2), (3, 2) (4, 1)] block groups as opposed to [(3, 2), 3, 2), (4, 1)], which results in an additional $4\times$ downsampling and 6 additional residual blocks. 
We used the same parameters to train the NeRF-VAE on both datasets: Models were trained for $2.5\times10^5$ iterations with batch size 192, learning rate of $5\times10^{-4}$ and Adam optimizer.
The training objective is to maximize the ELBO, which involves reconstructing an image with volumetric rendering. We sample 32 points from the coarse scene net, and we reuse those points plus sample additional 64 points for evaluating the fine scene net.
\citet{nerfvae} were able to set the integration cutoff $t_\text{far}=15$,
but for the data generated for this paper,
this value was too small and we used $t_\text{far}=40$.
This effectively reduces the number of points sampled per unit interval,
which might explain higher reconstruction and depth-estimation errors. 

\section{Datasets}
\label{secapp:data}
Here, we describe both of the 3D datasets used.

\subsection{CLEVR-3D}
\label{secapp:data-clevr}
To allow for maximum interoperability with the 2D version of CLEVR, we
use the scene metadata for the CLEVR dataset provided by \cite{multiobjectdatasets19}.
While this data does not specify the original (randomized) camera positions, we were able to recover
them based on the provided object coordinates in camera space via numerical root finding.
We were therefore able to exactly reconstruct the scenes in Blender, except for the light positions,
which were rerandomized.
To ensure a coherent background in all directions, we added a copy of the backdrop used by CLEVR,
facing in the opposite direction.
We rendered RGB-D views of the scene from the original camera positions
and two additional positions obtained by rotating the camera by $120^\circ$/$240^\circ$ around the
z axis. 
We test on the first $500$ scenes of the validation set, using RGB images from the original view points as input.

\subsection{MultiShapeNet}
\label{secapp:data-shapenet}
For the \shapenetname{} dataset, we start with the same camera, lighting, and background setup 
which was used for CLEVR-3D. For each scene, we first choose the number of objects uniformly
between 2 and 4. We then insert objects one by one. For each one, we first uniformly select
one of the \emph{chair}, \emph{table}, or \emph{cabinet} categories. We then uniformly choose a
shape out of the training set of shapes provided for that category in the ShapeNetV2 dataset \cite{shapenet2015},
leaving the possibility for future evaluations with unseen shapes. We insert the object at a random
$x,y$ position in $[-2.9, 2.9]^2$, scale its size by a factor of $2.9$, and choose its $z$ position
such that the bottom of the object is aligned with $z=0$. To prevent intersecting objects and reduce the
probability of major occlusions, we compute the radius $r$ of the newly inserted object's bounding circle
in the $xy$-plane. If a previously inserted object has radius $r'$, and their $xy$-distance
$d < 1.1 (r + r')$, we consider the objects to be too close, remove the last inserted object, and
sample a new one. If the desired number of objects has not been reached after $20$ iterations,
we remove all objects and start over.

\section{Model Parameters}
We report the hyperparameters used for \modelname{} in \cref{tab:hyperparams}. In addition to the
architectural parameters introduced above, we note the importance of the standard deviation of the
color distribution $\sigma_C$ (\cref{eq:color_dist}) for tuning the relative importance of
color compared to depth. We also report how we ramp up the overlap loss $\mathcal{L}_O$:
From the beginning of training to the iteration at which we start the ramp up, we set $k_O = 0$.
During the ramp up period, we linearly increase $k_O$ until it reaches the maximum value
$\hat{k}_O$.
We train using the Adam optimizer with default parameters, and an initial learning rate of
$4e-4$. We reduce the learning rate by a factor of $0.5$ every 100k iterations.
Finally, we note that for the 3D models, we used gradient norm clipping during training, \ie,
at each iteration, we determine the L2 norm of the gradients of our model as if they would form a single vector.
If this norm exceeds $1$, we divide all gradients by that norm. When training on \shapenetname{}, we find
that very rare, extremely strong gradients can derail training. We therefore skip all training steps with
norm greater than $1000$ entirely.
\label{secapp:hyperparameters}

\end{document}